\documentclass[]{arxiv_preprint}

\usepackage{amsmath}
\usepackage{amsfonts}
\usepackage{xspace}
\usepackage{wrapfig}

\newcommand{\ourmethod}{SG-WAM\xspace}

\title{\textcolor{metablue}{SG-WAM}: Text-grounded and Spatial-aware Semantic Guidance for World-Action Models}

\author[1,\ast]{Junjie He}
\author[1,\ast]{Junfeng Li}
\author[1,\ast,\dagger]{Zhide Zhong}
\author[1,\ast]{Haodong Yan}
\author[1]{Ruixin Li}
\author[1]{Yangyang Zheng}
\author[1]{Jiaguan Zhu}
\author[1]{Tianran Zhang}
\author[1]{Yuqiao Du}
\author[2]{Wen Chen}
\author[2]{Shunbo Zhou}
\author[1,\ddagger]{Haoang Li}

\affiliation[1]{The Hong Kong University of Science and Technology (Guangzhou), Guangzhou, China}
\affiliation[2]{Ola Dimensions, Shenzhen, China}
\contribution[\ast]{Equal contribution}
\contribution[\dagger]{Project Leader}
\contribution[\ddagger]{Corresponding author}

\hypersetup{
  pdftitle={SG-WAM: Text-grounded and Spatial-aware Semantic Guidance for World-Action Models},
  pdfauthor={Junjie He, Junfeng Li, Zhide Zhong, Haodong Yan, Ruixin Li, Yangyang Zheng, Jiaguan Zhu, Tianran Zhang, Yuqiao Du, Wen Chen, Shunbo Zhou, and Haoang Li}
}

\abstract{
  World-Action Models (WAMs) have emerged as a promising paradigm for robotic manipulation.
However, most existing WAMs generate future videos and actions by relying mainly on visual cues rather than language instructions, since off-the-shelf text encoders embed instructions independently of visual observations.
As a result, the videos predicted by these WAMs are often semantically misaligned with their corresponding language instructions, which degrades the accuracy of the predicted actions.
To overcome this limitation, we propose \ourmethod, a semantic guidance method for world-action models that leverages a vision-language model (VLM) as a semantic planner to enhance the instruction-grounding capacity of world-action models.
Specifically, we train a VLM-based planner to predict text-grounded and spatial-aware semantic foresight. 
The text-grounded semantic foresight grounds the instruction by identifying the correct target objects, and the spatial-aware semantic foresight provides the scene geometry for precise manipulation.
We then inject this foresight into the world-action model as high-level semantic guidance, ensuring that both future-video generation and action prediction faithfully follow the language instruction.
Extensive experiments in simulation and the real world demonstrate the superiority of our semantic guidance method, showcasing precise manipulation and strong instruction-following capabilities.

}

\metadata[Project Page]{\url{https://livfour.github.io/SG-WAM/}}
\metadata[Date]{August 2026}

\begin{document}

\maketitle

\section{Introduction}
\label{sec:intro}

World-Action Models (WAMs) have recently emerged as a promising paradigm for robotic manipulation~\cite{uwm,worldvla,dit4dit,maskwam,dswam,svam}.
Instead of mapping observations and instructions directly to actions as in vision-language-action (VLA) policies~\cite{brohan2023rt,zitkovich2023rt,kim2024openvla,BlackK-RSS-25,zhong2025flowvla,dualcot,song2025reconvla}, WAMs learn physical dynamics and task-relevant temporal structure by jointly predicting future observations and actions.
This predictive formulation leads to strong performance and generalization across a wide range of manipulation tasks, and further allows WAMs to be pretrained on large-scale video data in addition to robot demonstrations~\cite{unipi,gr1}.

Despite this progress, most existing video-based WAMs~\cite{vpp,lingbotva,genieenvisioner,fastwam} generate future frames and actions by relying mainly on visual cues rather than on the language instruction. 
The instruction enters generation only through text embeddings from off-the-shelf text encoders such as CLIP and T5~\cite{radford2021learning,raffel2020exploring}, computed independently of the visual observation.
The action expert therefore relies on dynamic features that are poorly aligned with the instruction~\cite{liberoplus}. This poor alignment directly degrades manipulation performance.

\clearpage
\begin{wrapfigure}{r}{0.5\textwidth}
    \centering
    \captionsetup{font=footnotesize}
    \includegraphics[width=\linewidth]{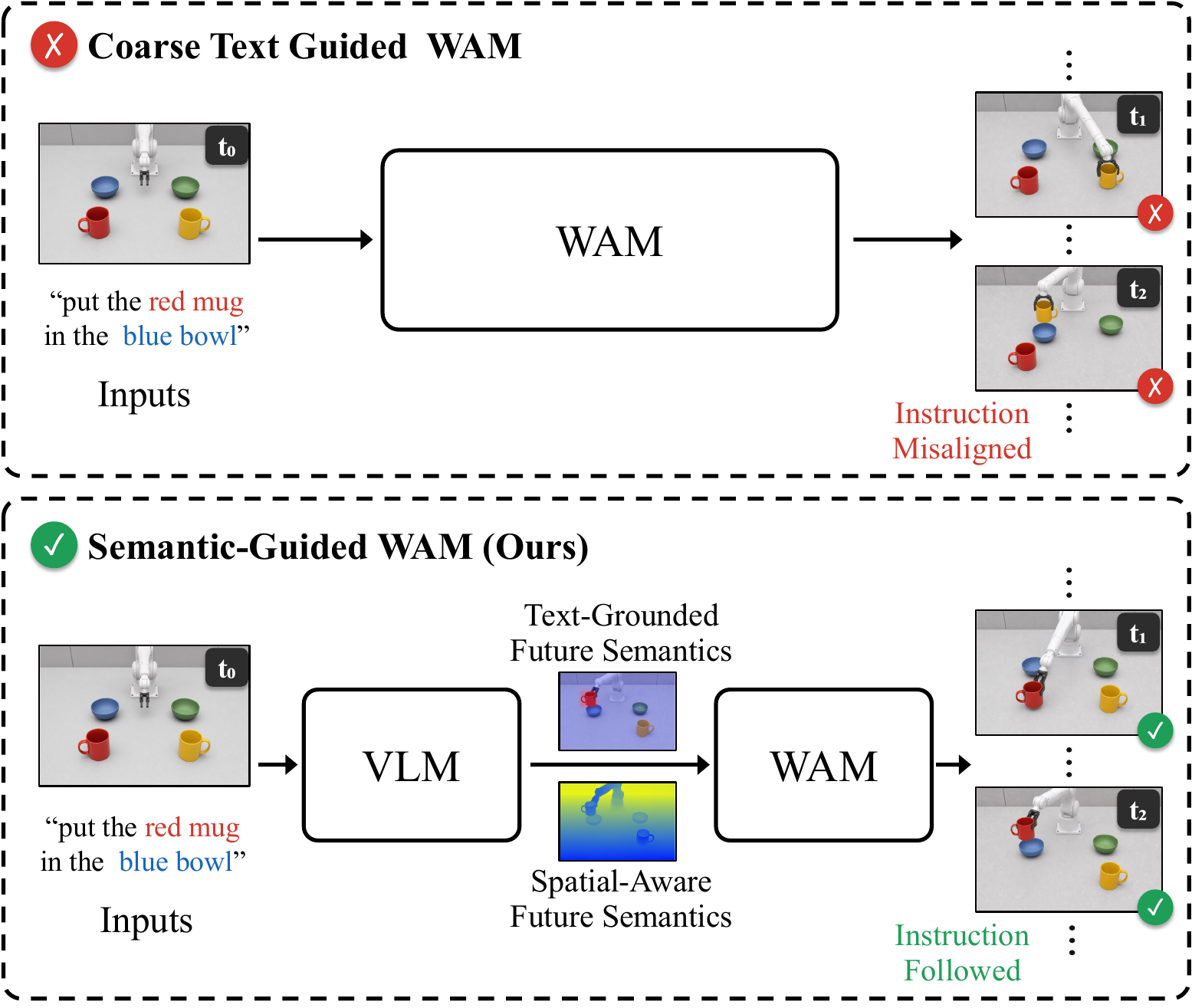}
    \caption{Our method introduces a VLM semantic planner that predicts text-grounded and spatial-aware semantic foresight of the intended future and uses it to guide the world-action model, so that both the generated videos and actions faithfully follow the language instruction.}
    \label{fig:teaser}
\end{wrapfigure}

To overcome this limitation, there have been efforts to use explicit spatial cues for world-action models~\cite{maskwam,maskwm}, but such cues still fall short of fine-grained and instruction-level semantics.
In this paper, we propose \ourmethod, a semantic guidance paradigm for world-action models (see Figure~\ref{fig:teaser}). Within this paradigm, world-action models are guided by text-grounded and spatial-aware semantic foresight instead of observation-independent text embeddings: the text-grounded semantics ground the instruction in the scene so that the model acts on the correct target objects, while the spatial-aware semantics supply the scene geometry needed to manipulate them precisely.
Together, they offer a fine-grained and structured blueprint for both future video and action prediction.

\begin{figure*}[t]
    \centering
    \includegraphics[width=\textwidth]{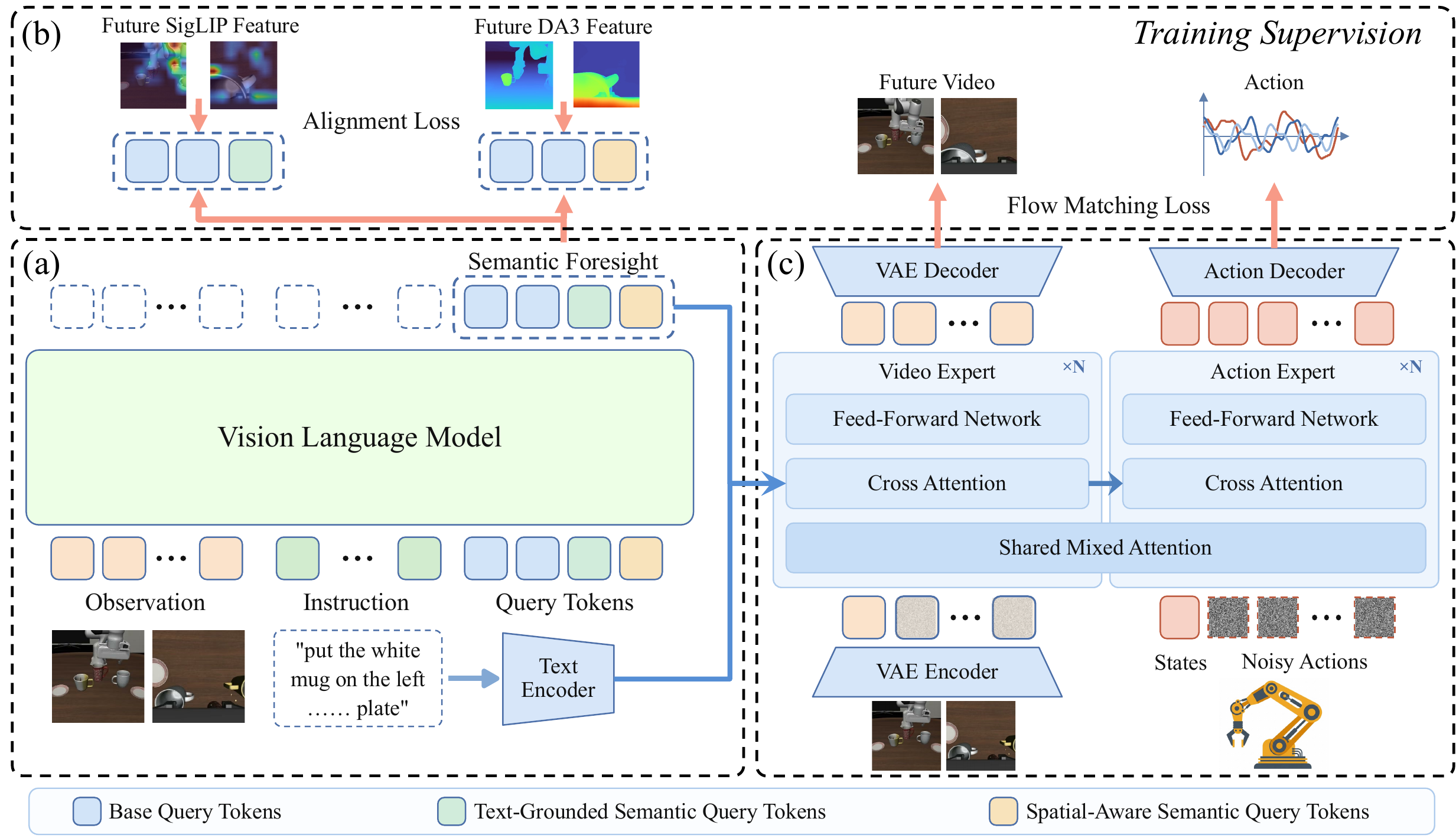}
    \caption{\textbf{Overview of our \ourmethod.} (a) From the current observation and the language instruction, the vision-language model (VLM) planner reads out text-grounded and spatial-aware semantic foresight from a base query group and two modality-specific query groups. (b) During training, the predicted foresight is aligned with the SigLIP2 and Depth Anything~3 features of the future keyframes, while the generated video and actions are supervised by a flow-matching loss. (c) Conditioned on this semantic guidance, the world-action model generates the future frames and the action chunk through a video expert and an action expert coupled by joint attention.}
    \label{fig:overview}
\end{figure*}

Specifically, as shown in Figure~\ref{fig:overview}, our \ourmethod consists of two components: a VLM planner that predicts the semantic guidance, and a world-action model that generates the future video and actions under this guidance.
We implement the planner by appending learnable query tokens to the image--instruction sequence and decoding the foresight from their hidden states in a single forward pass.
The tokens consist of a base group and two modality-specific groups: the base group learns the shared semantic knowledge, while the two specific groups learn text-grounded and spatial-aware semantics, respectively.
For the text-grounded semantic guidance, we align its output representations with the future-frame features of a language-aligned foundation model~\cite{siglip2}, which grounds the predicted semantics in the language instruction.
For the spatial-aware semantic guidance, we align its representations with those of a geometry foundation model~\cite{lin2025depth}, which provides the scene geometry required for precise manipulation.
The semantic foresight serves as a high-level blueprint that conditions the video expert, steering future prediction toward the intended semantics while preserving the pretrained generative ability of the video expert.
The action expert then decodes actions from this foresight-conditioned video expert, so that the predicted video and the executed actions stay consistent with the language instruction.

We comprehensively evaluate \ourmethod across both simulation benchmarks and a real-world robot platform.
Experiments on the LIBERO~\cite{liu2023libero} and LIBERO-Plus~\cite{liberoplus} benchmarks demonstrate that our method achieves state-of-the-art (SOTA) performance and remains robust under environmental and instruction perturbations.
Furthermore, real-world robot experiments confirm that our model transfers its instruction-following and precise-manipulation capabilities to physical environments.
In summary, our contributions are:
\begin{itemize}
    \item We propose a new semantic-guided prediction paradigm for world-action models, guiding both future and action prediction with text-grounded and spatial-aware semantic foresight rather than observation-independent text embeddings.

    \item We introduce \ourmethod, an instantiation of this paradigm built on a VLM planner that predicts complementary text-grounded and spatial-aware semantic foresight, enabling both future video and action generation to faithfully follow the language instruction.

    \item Extensive experiments in simulation and the real world show that \ourmethod achieves state-of-the-art performance, delivering precise manipulation and strong instruction following.
\end{itemize}

\section{Related Work}
\label{sec:related}

\paragraph{World-Action Models for Embodied Control}
World-action models leverage the generative priors of video backbones to predict how the scene evolves under interaction.
Early attempts generate instruction-conditioned videos and recover actions from them~\cite{unipi}, pretrain policies with video generation objectives~\cite{gr1,gr2,rynnvla}, or condition actions on the predictive representations of video diffusion models~\cite{vpp}.
More recent systems couple video foundation models with action experts at scale: Genie Envisioner~\cite{genieenvisioner} attaches a flow-matching action decoder to an instruction-conditioned video diffusion backbone, Cosmos Policy~\cite{cosmospolicy} directly fine-tunes a pretrained video model into a policy, LingBot-VA~\cite{lingbotva} interleaves video and action tokens for causal video-action world modeling, and a family of unified architectures jointly models the two modalities within a single transformer~\cite{uwm,uva,worldvla,dit4dit,fastwam}.
However, across these designs, the instruction enters generation only as coarse text embeddings~\cite{radford2021learning,raffel2020exploring} or plain text tokens, and the predicted futures capture physical dynamics rather than instruction semantics, so the generation easily drifts away from the task.

\paragraph{Guidance for Robot Manipulation}
Beyond plain text conditioning, many efforts supply manipulation models with intermediate guidance.
One line predicts visual subgoals: CoT-VLA~\cite{zhao2025cot} generates subgoal images as a visual chain of thought before decoding actions, and DreamVLA~\cite{zhangdreamvla} forecasts world knowledge such as dynamic regions and depth ahead of acting.
Within world-action models, MaskWAM~\cite{maskwam} takes first-frame masks as visual prompts and predicts future masks alongside RGB frames, Mask World Model~\cite{maskwm} forecasts the evolution of semantic masks instead of pixels, and OA-WAM~\cite{oawam} decomposes the scene into addressable object slots.
A third line plans with language: VLP~\cite{vlp} interleaves language subgoals with text-conditioned video generation, and DSWAM~\cite{dswam} invokes a VLM to decompose coarse commands into subtask text for the WAM executor.
However, pixel-level subgoals entangle task intent with appearance details, explicit masks and slots carry little instruction-relevant semantics beyond object layout, and subgoal text remains too coarse to specify where and when interactions should happen.
\ourmethod instead derives fine-grained text-grounded and spatial-aware foresight in feature space from a VLM planner and uses it to jointly steer future prediction and action decoding.

\section{Methodology}
\label{sec:method}

\subsection{Overview}
\label{sec:method_overview}
As shown in Figure~\ref{fig:overview}, the VLM planner takes the current observation $\mathbf{o}_t$ and the language
instruction $\ell$, together with a set of learnable query tokens.
From the hidden states of these query tokens, it predicts two dense feature maps for future keyframes: $\hat{\mathbf{F}}_{\mathrm{tg}}$ grounds the instruction on the task-relevant objects, and $\hat{\mathbf{F}}_{\mathrm{spa}}$ captures the
scene geometry required for precise manipulation.
During training, the two maps are supervised by frozen teacher encoders that
provide target features from future frames, driving the query states to
anticipate how the scene should evolve.

The world-action model, which couples a video expert with an action expert, generates the future video $\mathbf{o}_{t+1:t+T}$ of $T$ frames and the action chunk $\mathbf{a}_{t:t+H-1}$ of $H$ steps under the guidance of this foresight. We model the resulting conditional distribution as
\begin{equation}
    p_{\theta}\!\left(
        \mathbf{a}_{t:t+H-1},\,
        \mathbf{o}_{t+1:t+T}
        \,\middle|\,
        \mathbf{o}_t,\ell,\mathbf{s}_t,
        \hat{\mathbf{F}}_{\mathrm{tg}},
        \hat{\mathbf{F}}_{\mathrm{spa}}
    \right),
    \label{eq:wam_overview}
\end{equation}
where $\theta$ denotes the model parameters and $\mathbf{s}_t$ is the robot state that the action expert receives.

We train the model in three stages: we first train the semantic planner alone to
predict the semantic guidance of future keyframes, then co-train it with the
video expert, and finally add the action expert.
At inference, we discard the teacher encoders, and a single planner forward pass supplies
the foresight that guides generation.

\subsection{Semantic Foresight Learning}
\label{sec:method_planner}

We instantiate the semantic planner with Qwen3.5~\cite{qwen35}.
To predict future features in a single forward pass, we append learnable query
tokens to the image--instruction sequence and take their hidden states.
To capture the semantic knowledge shared by the text-grounded and spatial-aware
targets, we split the tokens into a base group and two modality-specific groups.
Both branches read the base group, while each modality-specific group captures
only what its own target requires.
Branch $m$ then reads the concatenated states
$\mathbf{H}_m=[\mathbf{H}_{\mathrm{base}};\mathbf{H}_m^{\mathrm{spec}}]$.
We reuse the same tokens for $K$ future keyframes at increasing offsets, so the
foresight covers a horizon instead of a single instant.
The planner is applied to each camera view independently and with shared
weights, so the same instruction is grounded separately in every view.
A lightweight resampler $\mathcal{R}_m$ then decodes each $\mathbf{H}_m$ into a
dense feature map.
To produce this map, a bank of $P$ learnable grid queries $\mathbf{U}_m$
cross-attends to $\mathbf{H}_m$ and the image states $\mathbf{E}$.
By attending to both sources, the resampler recovers fine spatial details from the observation while the query states specify how it should change:
\begin{equation}
    \hat{\mathbf{F}}_m
    =
    \mathcal{R}_m\!\left([\mathbf{E};\mathbf{H}_m],\,\mathbf{U}_m\right)
    \in\mathbb{R}^{P\times d_m},
    \label{eq:resampler}
\end{equation}
where $m\in\{\mathrm{tg},\mathrm{spa}\}$ indexes the two semantic branches.
Both maps use $P{=}256$ tokens to match the token layout of the two teachers, and the feature dimension $d_m$ is $1024$ for the
text-grounded map and $2048$ for the spatial-aware map.

Two frozen teachers supervise these predictions using features extracted from the actual future frames.
The text-grounded target $\mathbf{F}_{\mathrm{tg}}$ comes from the
penultimate-layer patch tokens of SigLIP2~\cite{siglip2}, which provides a language-aligned visual feature space with dense spatial structure.
The spatial-aware target $\mathbf{F}_{\mathrm{spa}}$ comes from the dense
features of Depth Anything~3~\cite{lin2025depth}, which encode object layout,
surface geometry, and relative distances.
We align to these features rather than directly regressing raw depth, so that geometry is represented in a dense-token format before both branches are projected into a common guidance space.
With both targets detached, the planner is optimized to match them:
\begin{equation}
    \mathcal{L}_{\mathrm{plan}}
    =
    \operatorname{MSE}(\hat{\mathbf{F}}_{\mathrm{tg}},\mathbf{F}_{\mathrm{tg}})
    +
    \lambda_{\mathrm{spa}}
    \operatorname{SL_1}(\hat{\mathbf{F}}_{\mathrm{spa}},\mathbf{F}_{\mathrm{spa}}),
    \label{eq:planner_loss}
\end{equation}
where $\operatorname{MSE}$ and $\operatorname{SL_1}$ denote the mean-squared and
smooth-$L_1$ losses, both averaged over tokens and feature channels,
$\lambda_{\mathrm{spa}}$ balances the two terms, and the loss is averaged over the
$K$ predicted keyframes.

\subsection{Semantic-Guided World-Action Model}
\label{sec:method_guidance}

We build the world-action model on a pretrained video diffusion model and couple
its video expert with an action expert through joint
attention~\cite{genieenvisioner,fastwam}.
To condition it on the predicted foresight, we fuse the text-grounded and
spatial-aware maps into a single guidance representation $\mathbf{Z}$ through a
learned gate:
\begin{equation}
    \mathbf{Z}
    =
    \operatorname{Proj}_{\mathrm{tg}}(\hat{\mathbf{F}}_{\mathrm{tg}})
    +
    \sigma(g)\,
    \operatorname{Proj}_{\mathrm{spa}}(\hat{\mathbf{F}}_{\mathrm{spa}}),
    \label{eq:fusion}
\end{equation}
where $\operatorname{Proj}_{\mathrm{tg}}$ and $\operatorname{Proj}_{\mathrm{spa}}$ are
linear projections into the guidance space, $\sigma$ is the sigmoid function, and
$g$ is a learnable scalar gate.
Rather than replacing the original text conditioning, we retain the original text cross-attention and inject $\mathbf{Z}$ into every block of the video expert through an additional parallel cross-attention branch. 
Each guidance token carries a positional encoding of its keyframe time and spatial
location, so it guides the region it describes.
To amplify the effect of the guidance at inference through classifier-free
guidance, we randomly drop the guidance during training so that the model also
learns to generate without it.
The action expert needs no direct injection, since it inherits the guidance
through its joint attention with the video expert.
Both experts are trained with a flow-matching objective.
For each expert, we form $\mathbf{x}_\tau=(1-\tau)\mathbf{y}+\tau\boldsymbol{\epsilon}$
from the clean target $\mathbf{y}$ and Gaussian noise $\boldsymbol{\epsilon}$ at a
sampled time $\tau\in[0,1]$, and regress the velocity
$\mathbf{u}=\boldsymbol{\epsilon}-\mathbf{y}$:
\begin{equation}
    \mathcal{L}_{\mathrm{gen}}
    =
    \underbrace{\mathbb{E}\big\|\hat{\mathbf{v}}_{\mathrm{vid}}-\mathbf{u}_{\mathrm{vid}}\big\|_2^2}_{\mathcal{L}_{\mathrm{vid}}}
    +
    \lambda_{\mathrm{act}}\,
    \underbrace{\mathbb{E}\big\|\hat{\mathbf{v}}_{\mathrm{act}}-\mathbf{u}_{\mathrm{act}}\big\|_2^2}_{\mathcal{L}_{\mathrm{act}}}.
    \label{eq:gen_loss}
\end{equation}
Here $\hat{\mathbf{v}}$ denotes the velocity predicted by each expert, the
subscripts $\mathrm{vid}$ and $\mathrm{act}$ refer to the video and action
experts, and $\lambda_{\mathrm{act}}$ balances the two terms.

\subsection{Three-Stage Training Paradigm}
\label{sec:method_training}

\paragraph{Stage 1: Semantic Planner Pre-training.}
From only the current frame and the instruction, the planner must reproduce the
dense features that the frozen teachers extract from the $K$ future keyframes.
We optimize the query-token embeddings and the two resampler heads jointly with
the VLM, keeping all teacher encoders frozen,
under the alignment loss $\mathcal{L}_{\mathrm{plan}}$ in
Eq.~\eqref{eq:planner_loss}.
After this stage, the semantic planner can predict text-grounded and
spatial-aware foresight from a single observation.

\paragraph{Stage 2: Semantic-Guided World Model Co-Training.}
We then initialize the world model from a pretrained video diffusion model and
connect it to the semantic planner through the gated fusion and cross-attention
described in Sec.~\ref{sec:method_guidance}, so that training starts from the
intact pretrained behavior.
The semantic planner and the world model are optimized jointly with
\begin{equation}
    \mathcal{L}_{\mathrm{stage2}}
    =
    \mathcal{L}_{\mathrm{vid}}
    +
    \lambda_{\mathrm{plan}}\,\mathcal{L}_{\mathrm{plan}},
    \label{eq:stage2_loss}
\end{equation}
where $\mathcal{L}_{\mathrm{vid}}$ is the video term of Eq.~\eqref{eq:gen_loss} and
$\lambda_{\mathrm{plan}}$ weights the alignment term, which keeps the planner
producing teacher-consistent foresight while adapting to generation.
This stage lets the world model learn to exploit the foresight before gradients from the action loss are introduced.

\paragraph{Stage 3: Semantic-Guided World-Action Co-Training.}
Finally, we attach the action expert and optimize the planner, the world model,
and the action expert together with
\begin{equation}
    \mathcal{L}
    =
    \mathcal{L}_{\mathrm{gen}}
    +
    \lambda_{\mathrm{plan}}\,\mathcal{L}_{\mathrm{plan}}.
    \label{eq:total_loss}
\end{equation}
Crucially, all components are conditioned on the planner's own predictions
rather than the teacher features, which matches deployment.

\section{Experiments}
\label{sec:exp}

\begin{table*}[!t]
\centering
\begingroup
\small
\setlength{\tabcolsep}{4pt}
\begin{tabular*}{\textwidth}{@{\extracolsep{\fill}}lcccccc}
\toprule
Method & Type & Spatial & Object & Goal & Long & Avg \\
\midrule
WorldVLA~\cite{worldvla} & WAM & 87.6 & 96.2 & 83.4 & 60.0 & 81.8 \\
GR00T-N1~\cite{bjorck2025gr00t} & VLA & 94.4 & 97.6 & 93.0 & 90.6 & 93.9 \\
$\pi_0$~\cite{BlackK-RSS-25} & VLA & 96.8 & 98.8 & 95.8 & 85.2 & 94.1 \\
GE-Act~\cite{genieenvisioner} & WAM & 98.2 & 97.6 & 95.8 & 94.4 & 96.5 \\
$\pi_{0.5}$~\cite{black2025pi} & VLA & \underline{98.6} & 98.2 & \underline{98.0} & 92.4 & 96.8 \\
OpenVLA-OFT~\cite{KimM1-RSS-25} & VLA & 97.6 & 98.4 & 97.9 & 94.5 & 97.1 \\
FastWAM~\cite{fastwam} & WAM & 98.2 & \textbf{100.0} & 97.0 & 95.2 & 97.6 \\
Motus~\cite{bi2025motus} & WAM & 96.8 & \underline{99.8} & 96.6 & 97.6 & 97.7 \\
LingBot-VA~\cite{lingbotva} & WAM & 98.5 & 99.6 & 97.2 & \textbf{98.5} & \underline{98.5} \\
\midrule
\ourmethod (Ours) & WAM & \textbf{99.4} & 99.4 & \textbf{98.2} & \underline{97.8} & \textbf{98.7}\\
\bottomrule
\end{tabular*}
\endgroup
\caption{
Results on the LIBERO benchmark \cite{liu2023libero}.
We report per-suite success rates on the Spatial, Object, Goal, and Long suites, along with their average.
``Type'' distinguishes vision-language-action models (VLA) from world-action models (WAM).
Bold indicates the best result, and underline indicates the second best.
}
\label{tab:main_results}
\end{table*}

\begin{table*}[!t]
\centering
\begingroup
\small
\setlength{\tabcolsep}{4pt}
\begin{tabular*}{\textwidth}{@{\extracolsep{\fill}}lcccccccc}
\toprule
Method & Layout & Camera & Init & Language & Light & BG & Noise & Avg \\
\midrule
OpenVLA~\cite{kim2024openvla} & 28.5 & 0.8 & 3.5 & 23.0 & 8.1 & 34.8 & 15.2 & 15.6 \\
WorldVLA~\cite{worldvla} & 38.0 & 0.1 & 27.9 & 41.6 & 43.7 & 17.1 & 10.9 & 25.0 \\
NORA~\cite{nora} & 62.1 & 2.2 & 37.0 & 65.1 & 45.7 & 58.6 & 12.8 & 39.0 \\
UniVLA~\cite{univla} & 31.9 & 1.8 & 46.2 & 69.6 & 69.0 & 81.0 & 21.2 & 43.9 \\
FastWAM~\cite{fastwam} & 60.7 & 16.4 & 44.5 & 68.9 & 78.2 & 53.7 & 37.7 & 50.0 \\
$\pi_0$~\cite{BlackK-RSS-25} & 68.9 & 13.8 & 6.0 & 58.8 & 85.0 & 81.4 & 79.0 & 53.6 \\
$\pi_0$-FAST~\cite{pi0fast} & 68.8 & \textbf{65.1} & 21.6 & 61.0 & 73.2 & 73.2 & 74.4 & 61.6 \\
OpenVLA-OFT~\cite{KimM1-RSS-25} & 74.2 & \underline{56.4} & 31.9 & \underline{79.5} & 88.7 & \textbf{93.3} & 75.8 & 69.6 \\
GE-Act~\cite{genieenvisioner} & \underline{83.0} & {49.6} & \underline{77.2} & 79.4 & \underline{94.8} & 85.1 & \underline{83.0} & \underline{77.8} \\
\midrule
\ourmethod (Ours) & \textbf{85.3} & 54.8 & \textbf{78.3} & \textbf{81.7} & \textbf{96.5} & \underline{88.9} & \textbf{90.7} & \textbf{81.3}\\
\bottomrule
\end{tabular*}
\endgroup
\caption{
Results on LIBERO-Plus~\cite{liberoplus}.
Each column perturbs one factor of the original LIBERO tasks, and Avg is the success rate over all perturbed tasks.
``Init'' and ``BG'' denote the robot initial state and the background texture, respectively.
Bold indicates the best result, and underline indicates the second best.
}
\label{tab:liberoplus}
\end{table*}

We comprehensively evaluate \ourmethod across both simulation benchmarks and a real-world robot platform.
We organize the experiments around four questions:

\begin{itemize}
    \item \textbf{Q1:} Does \ourmethod achieve state-of-the-art performance on standard manipulation benchmarks and on a real robot?
    \item \textbf{Q2:} Does semantic foresight improve robustness when the environment or the instruction is perturbed?
    \item \textbf{Q3:} Does the predicted foresight actually steer future videos and actions generation to follow the instruction?
    \item \textbf{Q4:} How much does each component of \ourmethod contribute to the final performance?
\end{itemize}

\subsection{Experimental Setup}
\label{sec:exp_setup}

\paragraph{Benchmarks.}
We comprehensively evaluate \ourmethod on three benchmarks:
\begin{itemize}
    \item \textbf{LIBERO}~\cite{liu2023libero} evaluates tabletop manipulation on four suites (Spatial, Object, Goal, and Long).
    Following the standard protocol~\cite{kim2024openvla}, we evaluate 500 trials per suite and report the success rate.
    \item \textbf{LIBERO-Plus}~\cite{liberoplus} extends LIBERO with seven perturbation factors, covering object layout, camera viewpoint, robot initial state, language rephrasing, lighting, background texture, and sensor noise.
    \item \textbf{Real-world experiments} are conducted on an AgileX Cobot dual-arm platform, with the setup and tasks described in Sec.~\ref{sec:exp_realworld}.
\end{itemize}

\begin{figure*}[t]
    \centering
    \includegraphics[width=\textwidth]{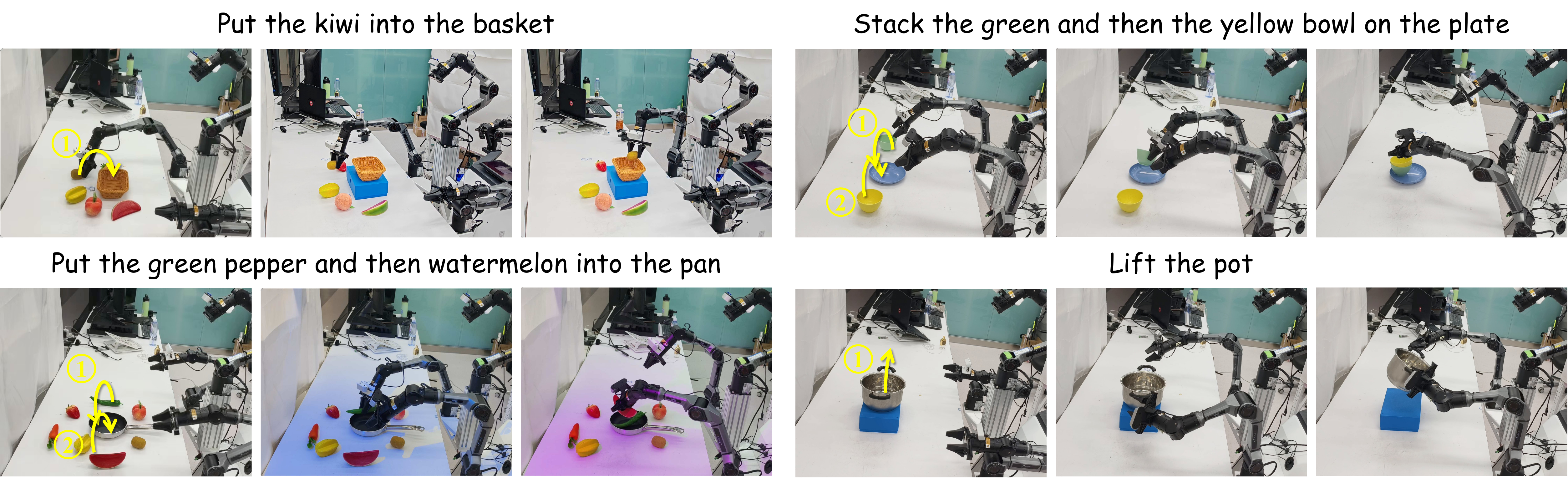}
    \caption{\textbf{Real-world qualitative results.} Qualitative execution sequences using our \ourmethod{} across four manipulation tasks, validating different generalization and manipulation capabilities: height generalization, stacking in the instructed order, target selection under distractors and lighting variation, and bimanual collaboration.}
    \label{fig:real_rollout}
\end{figure*}

\begin{wrapfigure}{r}{0.5\textwidth}
    \centering
    \captionsetup{font=footnotesize}
    \includegraphics[width=\linewidth]{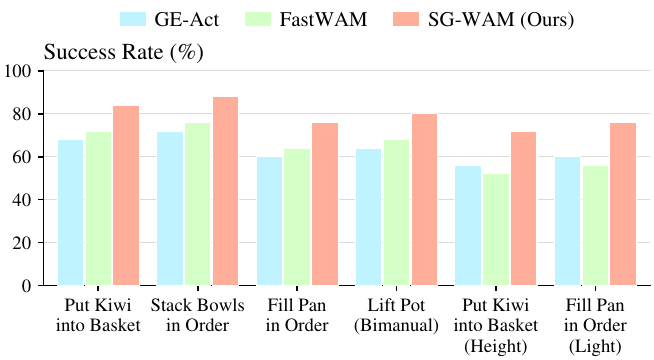}
    \caption{\textbf{Real-world multi-task results.} We report success rates (SR) across four different tasks as well as two generalization experiments.  ``Height'' evaluates kiwi placement with the basket at a height unseen during data collection, while ``Light'' evaluates ordered pan filling under unseen lighting conditions.}
    \label{fig:realworld}
\end{wrapfigure}

\paragraph{Baselines.}
We compare against two groups of methods (see Table~\ref{tab:main_results}).
\textbf{VLA models}, including OpenVLA-OFT~\cite{KimM1-RSS-25}, $\pi_0$ and $\pi_{0.5}$~\cite{BlackK-RSS-25,black2025pi}, and GR00T-N1~\cite{bjorck2025gr00t}, map observations and instructions directly to actions without modeling future dynamics.
\textbf{WAM models}, including WorldVLA~\cite{worldvla}, GE-Act~\cite{genieenvisioner}, Motus~\cite{bi2025motus}, FastWAM~\cite{fastwam}, and LingBot-VA~\cite{lingbotva}, jointly predict future observations and actions but condition generation only on coarse text embeddings.

\paragraph{Implementation Details.}
Our \ourmethod utilizes Qwen3.5 2B~\cite{qwen35} as the semantic planner.
The two representation spaces we use are SigLIP2~\cite{siglip2} for text-grounded semantics and Depth Anything~3~\cite{lin2025depth} for spatial-aware semantics.
We align to the penultimate-layer patch tokens of SigLIP2 and to the last-layer features of Depth Anything~3. 
The planner predicts $K{=}4$ future keyframes at evenly spaced offsets, using $32$ shared base tokens together with $32$ query tokens per semantic branch.
The world model is initialized from a pretrained LTX-Video~\cite{ltxvideo} diffusion model; the action expert is coupled to the video expert by joint attention as described in Sec.~\ref{sec:method_guidance}.
The model generates $T{=}9$ future frames and predicts an action chunk of $H{=}36$ steps via a flow-matching paradigm.
We conduct all three training stages on eight H100 GPUs.
Stage 1 runs for 20k steps with a learning rate of $4.2{\times}10^{-5}$ for the VLM and $4.2{\times}10^{-4}$ for the two resampler heads, weight decay $0.01$, and a cosine schedule with 2.5k warmup steps, where the spatial-aware term is weighted by $\lambda_{\mathrm{spa}}{=}4{\times}10^{-3}$.
Stages 2 and 3 each run for 30k steps with 1k warmup steps, using $2{\times}10^{-5}$ for the video expert, $1{\times}10^{-4}$ for the action expert, and $5{\times}10^{-5}$ for the semantic adapter, and retain the alignment term with $\lambda_{\mathrm{plan}}{=}0.25$ and the action term with $\lambda_{\mathrm{act}}{=}1.0$.
The semantic guidance is dropped with probability $0.15$ during training to enable classifier-free guidance at inference.

\subsection{Comparisons with State-of-the-Art Methods (Q1 \& Q2)}
\label{sec:main_results}

\paragraph{Results on LIBERO.}
As shown in Table~\ref{tab:main_results}, \ourmethod achieves the highest average success rate of 98.7\% on LIBERO, surpassing the strongest baseline LingBot-VA (98.5\%).
For a direct comparison with the two baselines used in our real-world experiments, \ourmethod outperforms FastWAM and GE-Act by 1.1 and 2.2 points, respectively.

\paragraph{Robustness Under Perturbations.}
LIBERO-Plus perturbs one factor of the environment at a time (Table~\ref{tab:liberoplus}).
\ourmethod consistently outperforms competing baselines, reaching the highest average success rate of 81.3\% ($+3.5$ points), with a notable improvement under language perturbation (81.7\%, $+2.2$ points).
This gain highlights the advantage of grounding the instruction in the current observation.
While existing WAMs embed the instruction independently of the visual observation, \ourmethod predicts its semantic foresight from both the instruction and the current scene to yield guidance that stays grounded when the appearance or wording changes.

\begin{figure*}[!t]
    \centering
    \includegraphics[width=\textwidth]{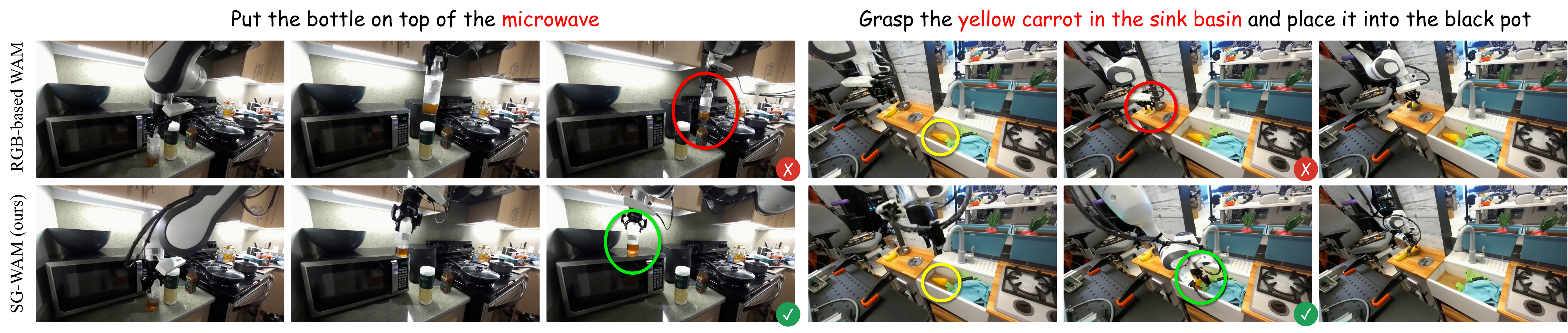}
    \caption{\textbf{Instruction-following comparison on DROID~\cite{droid}.} Videos generated from the same reference frame under out-of-distribution instructions, with the instructed target highlighted in red. For each instruction, the top row shows the RGB-based WAM and the bottom row shows \ourmethod, and red circles indicate the object or location acted upon.}
    \label{fig:if_vis}
\end{figure*}

\subsection{Multi-task Experiments in the Real World (Q1 \& Q2)}
\label{sec:exp_realworld}

\begin{wrapfigure}{r}{0.5\textwidth}
    \centering
    \captionsetup{font=footnotesize}
    \includegraphics[width=\linewidth]{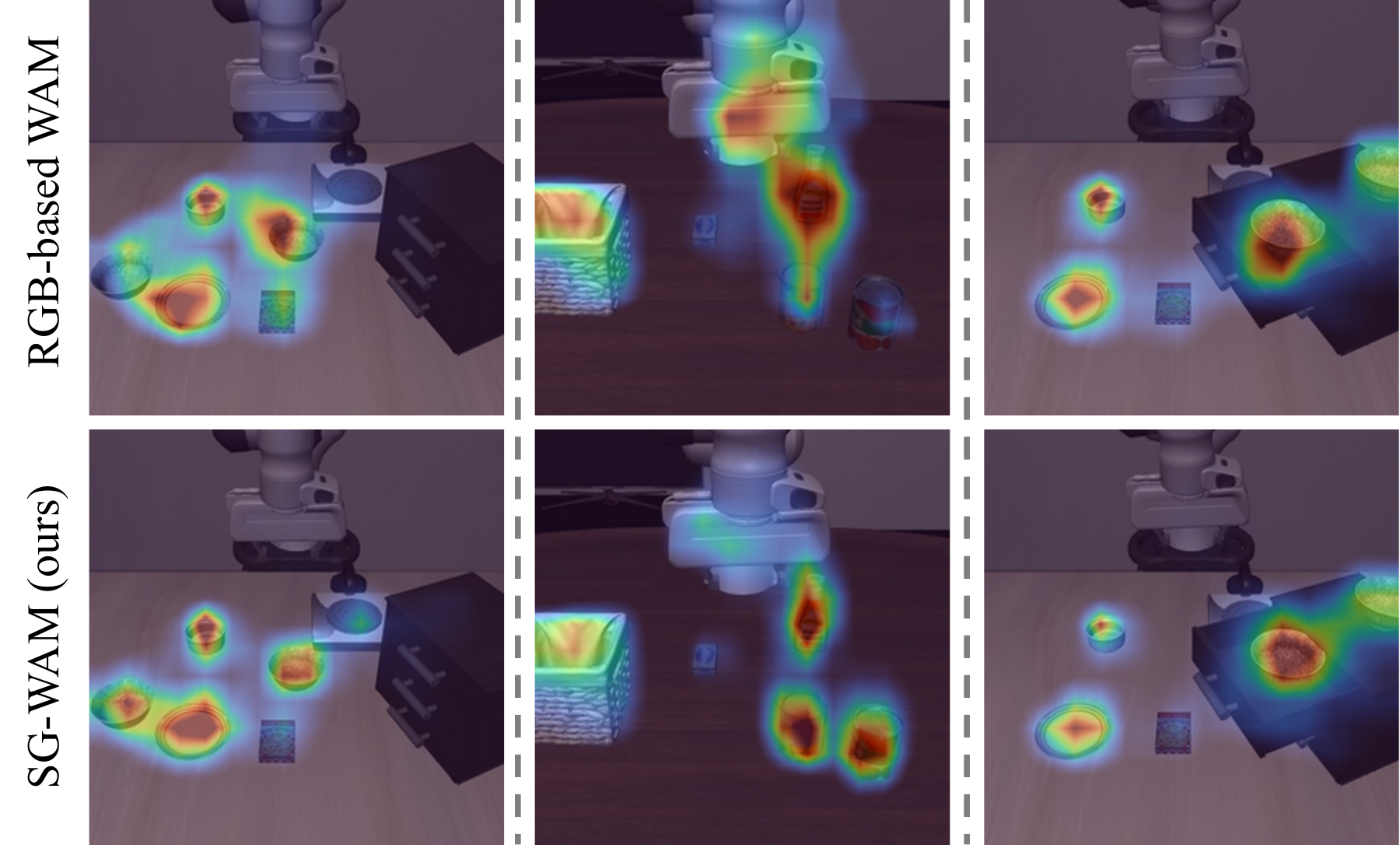}
    \caption{\textbf{Qualitative comparison of attention maps on LIBERO~\cite{liu2023libero}.} Action-to-video attention to the instruction-relevant object. The RGB-based WAM (top) spreads its attention over several objects and over the robot arm itself, while \ourmethod (bottom) concentrates it on the manipulated objects.}
    \label{fig:foresight_vis}
\end{wrapfigure}

\paragraph{Setup.}
We conduct real-world experiments on an AgileX Cobot dual-arm platform with two 6-DoF arms and parallel grippers, using one front-facing RGB camera as Eye-on-Base and one RGB camera on each wrist as Eye-on-Hand for visual inputs.

\paragraph{Tasks.}
We select four representative tasks (see Figure~\ref{fig:real_rollout}): Put kiwi into basket, Stack bowls in order, Fill pan in order, and Lift pot.
Each task involves multiple distractor objects and randomized initial layouts, and the pick-and-place tasks specify both the target objects and the order in which they are placed.
We collect 100 teleoperated demonstrations per task, and each model is tested on each task with 50 trials, using the success rate as the performance metric.
For generalization, we further repeat the basket task at a container height unseen in the demonstrations and the pan task under lighting conditions that differ from data collection, again with 50 trials each.

\paragraph{Results.}
As shown in Figure~\ref{fig:realworld}, \ourmethod consistently outperforms GE-Act and FastWAM across the four standard tasks, achieving the highest success rate in each case.
The gain is largest on the tasks in which the instructed objects have to be selected among distractors and placed in the commanded order.
Notably, all policies are trained only on demonstrations collected in the standard setting, yet \ourmethod remains effective when the container is raised to a height never seen during data collection.
\ourmethod also attains the best success rate in both generalization settings, and its accuracy under the shifted lighting matches that of the standard pan task.
This result suggests that foresight represented in semantic feature space remains reliable under shifts in scene geometry or appearance.

\subsection{Instruction-Following Analysis (Q3)}
\label{sec:exp_instruction}

\begingroup
\setlength{\emergencystretch}{2em}
We next examine whether the foresight is what makes the generated futures and actions follow the instruction, using two qualitative studies.

\paragraph{Out-of-Distribution Instruction Following.}
We train both the RGB-based WAM and \ourmethod on DROID~\cite{droid}, and let them generate future videos from the same reference frame under instructions unseen during training.
As shown in Figure~\ref{fig:if_vis}, the RGB-based WAM places the bottle on the stove instead of the microwave, and reaches for the wrong item instead of the carrot in the sink basin, while \ourmethod acts on the commanded object in both cases.
The RGB-based WAM receives an instruction embedding computed independently of the observation, and therefore relies mainly on visual cues, drifting toward the most salient object when the wording is unfamiliar.

\paragraph{Instruction-Grounded Attention Analysis.}
As shown in Figure~\ref{fig:foresight_vis}, we compare the action-to-video attention heatmaps of \ourmethod against the RGB-based WAM.
The RGB-based WAM spreads its attention over several objects and over the robot arm itself, whereas \ourmethod concentrates it on the object named in the instruction, indicating that the guidance directs the capacity of the model to the task-relevant region.
\endgroup

\subsection{Ablation Study (Q4)}
\label{sec:exp_ablation}

To evaluate the individual contributions of the semantic guidance, the dual-semantic design, and the three-stage training, we train ablated variants on the same data setting with identical hyperparameters and evaluate them on LIBERO.
The results are summarized in Table~\ref{tab:ablation}.

\begin{wraptable}{r}{0.5\textwidth}
\centering
\captionsetup{font=footnotesize}
\caption{
Ablation study on the core design choices of \ourmethod.
We report the average success rate (\%) on LIBERO, where each row removes or replaces one component.
Bold indicates the best result.
}
\label{tab:ablation}
\begingroup
\footnotesize
\setlength{\tabcolsep}{2pt}
\begin{tabular*}{\linewidth}{@{\extracolsep{\fill}}lc}
\toprule
Variant & Success Rate (\%) \\
\midrule
\ourmethod (full) & \textbf{98.7} \\
\midrule
\multicolumn{2}{l}{\textit{Semantic guidance}} \\
\quad w/o semantic guidance & 97.2 \\
\multicolumn{2}{l}{\textit{Dual-semantics design}} \\
\quad w/o spatial-aware semantics & 98.3 \\
\quad w/o text-grounded semantics & 97.7 \\
\quad w/o shared semantic query tokens & 98.4 \\
\multicolumn{2}{l}{\textit{Three-stage training}} \\
\quad single-stage training & 97.9 \\
\bottomrule
\end{tabular*}
\endgroup
\end{wraptable}

\paragraph{Semantic Guidance.}
Removing the semantic guidance and conditioning the backbone on text alone causes the largest drop. This supports the benefit of instruction-grounded guidance for joint video-action generation.

\paragraph{Dual-Semantics Design.}
Removing either semantics degrades performance. The removal of text-grounded semantics causes a noticeable drop, since it decides which object to act on. The removal of spatial-aware semantics also leads to a drop, as it provides the geometry for manipulating that object.
This evidence supports the benefit of text-grounded and spatial-aware semantic guidance.
Replacing the shared query tokens with specific query tokens causes a small but consistent drop.

\paragraph{Three-Stage Training.}
Training the planner, the world model, and the action expert jointly from scratch also degrades performance.
The action expert then starts learning before the planner produces reliable guidance, which supports training the three components in stages.

\section{Conclusion}
\label{sec:conclusion}

We proposed a novel paradigm that guides world-action models with text-grounded and spatial-aware semantic foresight rather than observation-independent text embeddings, and instantiate this paradigm in \ourmethod.
Our key insight is that a VLM planner can translate the instruction into a text-grounded and spatial-aware semantic foresight of the intended future, providing instruction-grounding and spatial-aware cues that are often missing from existing world-action models.
\ourmethod achieves state-of-the-art performance on LIBERO, the highest average robustness on LIBERO-Plus, and the best real-world success rates among the evaluated methods, with larger gains in the generalization tests.
These results suggest that text-grounded and spatial-aware foresight can effectively connect the reasoning capabilities of VLMs with the generative capabilities of world-action models.

\clearpage
\bibliographystyle{assets/plainnat}
\bibliography{references}

\end{document}